\documentclass[conference]{IEEEtran}
\IEEEoverridecommandlockouts

\usepackage{cite}
\usepackage{amsmath,amssymb,amsfonts}
\usepackage{algorithm}
\usepackage{algpseudocode}
\usepackage{graphicx}
\usepackage{textcomp}
\usepackage{xcolor}
\usepackage{comment}

\usepackage{hyperref}

\usepackage{multirow}
\def\BibTeX{{\rm B\kern-.05em{\sc i\kern-.025em b}\kern-.08em
    T\kern-.1667em\lower.7ex\hbox{E}\kern-.125emX}}
\begin{document}

\title{DASViT: Differentiable Architecture Search for Vision Transformer\\
}

\author{
\begin{minipage}[t]{0.3\textwidth}
\centering
Pengjin Wu \\
\textit{School of Computer Science} \\
\textit{and Electronic Engineering} \\
\textit{University of Surrey} \\
Guildford, UK \\
\href{mailto:pengjin.wu@surrey.ac.uk}{pengjin.wu@surrey.ac.uk}
\end{minipage}
\hfill
\begin{minipage}[t]{0.33\textwidth}
\centering
Ferrante Neri \\
\textit{School of Computer Science} \\
\textit{and Electronic Engineering} \\
\textit{University of Surrey} \\
Guildford, UK \\
\href{mailto:f.neri@surrey.ac.uk}{f.neri@surrey.ac.uk}
\end{minipage}
\hfill
\begin{minipage}[t]{0.3\textwidth}
\centering
Zhenhua Feng \\
\textit{School of Artificial Intelligence} \\
\textit{and Computer Science} \\
\textit{Jiangnan University} \\
Wuxi, China \\
\href{mailto:fengzhenhua@jiangnan.edu.cn}{fengzhenhua@jiangnan.edu.cn}
\end{minipage}
}

\maketitle

\begin{abstract}
Designing effective neural networks is a cornerstone of deep learning, and Neural Architecture Search (NAS) has emerged as a powerful tool for automating this process. Among the existing NAS approaches, Differentiable Architecture Search (DARTS) has gained prominence for its efficiency and ease of use, inspiring numerous advancements. Since the rise of Vision Transformers (ViT), researchers have applied NAS to explore ViT architectures, often focusing on macro-level search spaces and relying on discrete methods like evolutionary algorithms. While these methods ensure reliability, they face challenges in discovering innovative architectural designs, demand extensive computational resources, and are time-intensive. To address these limitations, we introduce Differentiable Architecture Search for Vision Transformer (DASViT), which bridges the gap in differentiable search for ViTs and uncovers novel designs. Experiments show that DASViT delivers architectures that break traditional Transformer encoder designs, outperform ViT-B/16 on multiple datasets, and achieve superior efficiency with fewer parameters and FLOPs.
\end{abstract}

\begin{IEEEkeywords}
Neural Architecture Search, Differentiable Architecture Search, Vision Transformer, Image Classification.
\end{IEEEkeywords}

\section{Introduction}
Neural Architecture Search (NAS), a cornerstone of Automated Machine Learning (AutoML), automates the design of network architectures tailored to specific tasks and datasets, reducing reliance on manual trial-and-error and expert intervention~\cite{elsken2019neural}. NAS defines a search space of potential architectures and employs advanced algorithms to identify optimal or near-optimal designs. Research has shown that NAS-derived architectures save time and effort and often surpass handcrafted networks in performance.

Early NAS methods, such as reinforcement learning and evolutionary algorithms, are effective but computationally expensive, requiring significant time and resources. The introduction of gradient-based methods, notably Differentiable Architecture Search (DARTS)~\cite{liu2018darts}, marks a significant advancement by reformulating the discrete search space into a continuous one and expressing operations (e.g., convolutions, pooling, skip connections) as weighted combinations. By treating architecture selection as a continuous optimisation problem, DARTS enables gradient descent to be directly applied to architecture parameters, dramatically reducing search costs and improving efficiency. This innovation establishes DARTS as a benchmark in NAS research.

Despite its effectiveness, DARTS faces several challenges in practical scenarios. First, maintaining and computing all candidate operations on each edge consumes substantial memory \cite{zhao2021memory}, which becomes problematic as the search space expands. Second, parameter-free operations, with their lower impedance to gradient flow, are disproportionately favoured, leading to an overreliance on simplistic structures that hinder the network’s representational capacity~\cite{xue2024improved}. Last, softmax normalization can give an unfair advantage to operations with initially higher weights, restricting exploration and potentially leading to instability and premature convergence to suboptimal solutions.

While much-existing research focuses on architecture search for conventional networks (e.g., CNNs, RNNs), the success of Transformers~\cite{vaswani2017attention} in NLP, combined with advances in computational hardware, has led to their increasing use in diverse tasks. In computer vision, Vision Transformers (ViT)~\cite{dosovitskiy2020image} have reshaped the paradigms for image classification, object detection, segmentation, etc. However, most NAS research on ViT relies on discrete evolutionary algorithms and typically searches only network parameters, not the structural design of Transformer encoders.

\begin{figure}[!t]
\centering
\includegraphics[trim={0 0 120mm 90mm}, clip, width=1\linewidth]{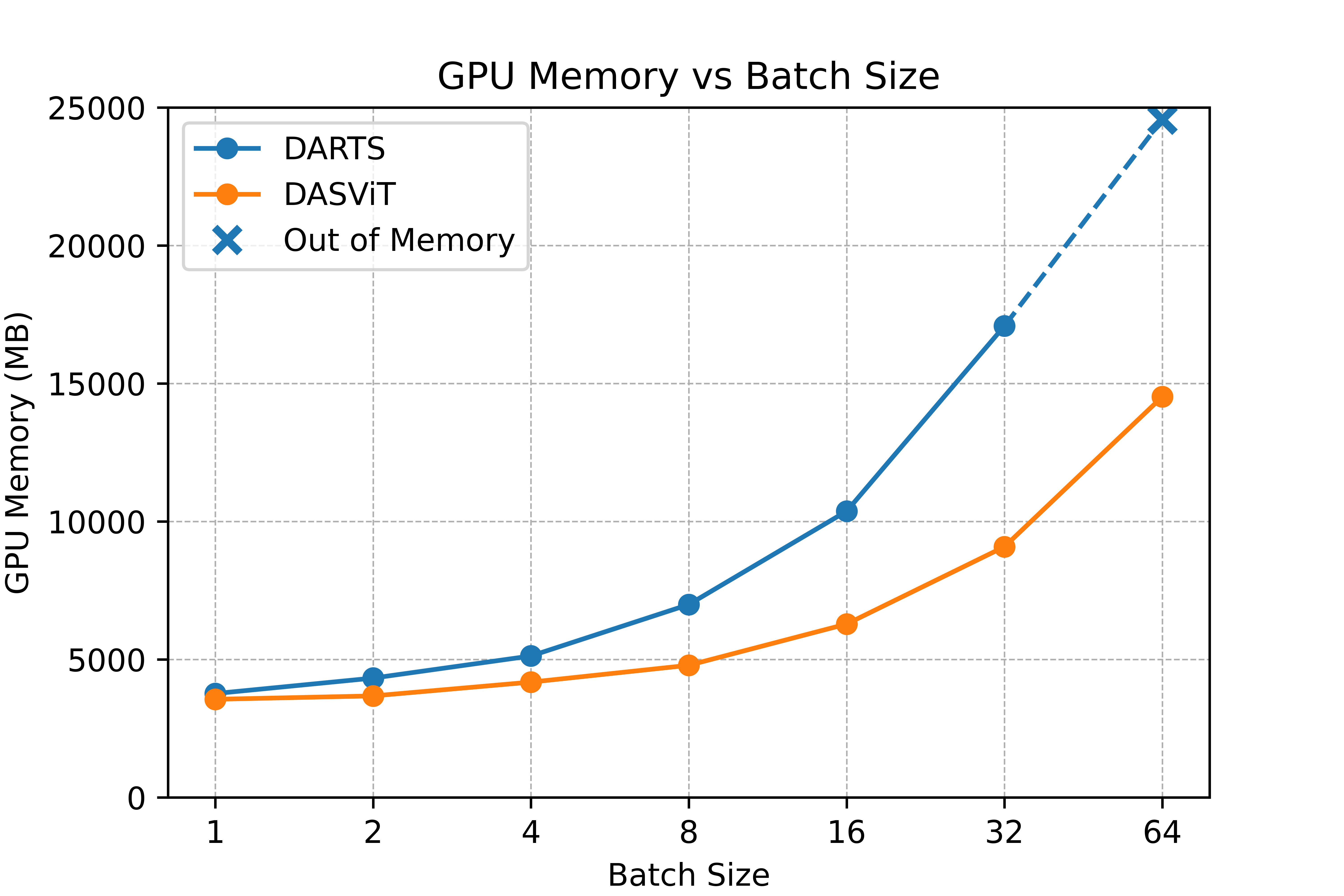}
\caption{Comparison of memory usage between applying DARTS directly to Vision Transformer and our proposed DASViT. Experiments were conducted on an Nvidia GeForce RTX 3090 card with the input image size of 224$\times$224, patch size of 16$\times$16, and embedding dimension of 768.}
\label{fig1}
\end{figure}

\begin{figure*}[!t]
\centering
\includegraphics[width=\linewidth]{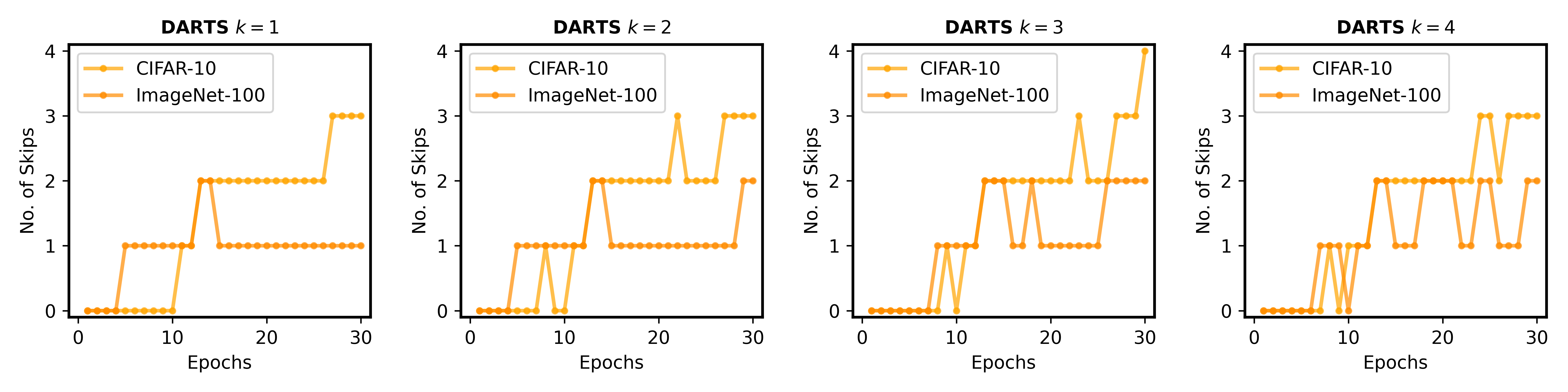}
\caption{The number of \texttt{skip-connect} operations (\texttt{identity}) consistently increases and dominates during independent DARTS searches (run \textit{k} = 4 times) conducted on CIFAR-10 and ImageNet-100.}
\label{fig2}
\end{figure*}

\begin{figure*}[!t]
\centering
\includegraphics[trim={0 0 200mm 80mm}, clip, width=.95\linewidth]{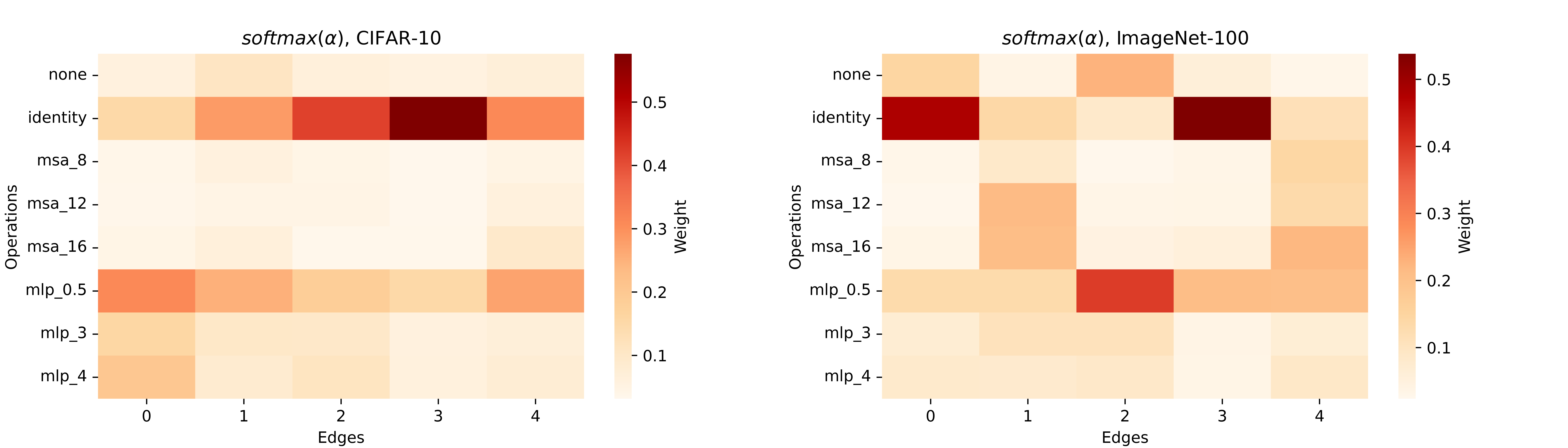}
\caption{This heatmap illustrates the distribution of operation weights ($\mathit{softmax}(\alpha)$) at the 30th epoch during the DARTS search on the CIFAR-10 (left) and ImageNet-100 (right) datasets. It can be observed that the weights predominantly concentrate on the \texttt{skip-connect} operation (\texttt{identity}). On certain edges, the weights also tend to cluster around Multi-Layer Perceptron (MLP) or Multi-head Self-Attention (MSA) operations, which are associated with different datasets' characteristics.}
\label{fig3}
\end{figure*}

In light of the above shortcomings, we attempt to build on the ideas of DARTS by representing the Transformer encoder as a Directed Acyclic Graph (DAG) and employing a differentiable method to search for operations in a continuous space. We aim to discover novel Transformer encoder structures. However, when applying DARTS to Vision Transformers (with the specific operation set defined in Section~\ref{subsec:search_space}), the inherent complexity and computational overhead of Transformer models amplify various issues. As shown in Figure~\ref{fig1}, even under fixed embedding dimensions, input image size, and patch size, merely increasing the batch size results in GPU memory (24 GB VRAM) exhaustion. Meanwhile, as illustrated in Figure~\ref{fig2} and Figure~\ref{fig3}, parameter-free operations (e.g. skip-connect) gain progressively higher weights during the search process, leading to final architectures dominated by skip-connect. On simpler datasets (such as CIFAR-10), the final architecture may even lack Multi-head Self-Attention (MSA) blocks entirely. This suggests that directly applying DARTS to Vision Transformers exacerbates issues related to memory consumption and the unfair selection of operations induced by softmax, necessitating targeted solutions.

To address these challenges, we propose a novel architecture search algorithm named DASViT. Using ViT-B/16~\cite{dosovitskiy2020image} as a baseline, we reformulate the problem into a differentiable continuous space, enabling the use of gradient-based optimisation methods to search for pure self-attention-driven Vision Transformer models. The design objectives of DASViT are to fully exploit the efficiency and scalability of DARTS, as demonstrated in CNN-based architecture search while addressing the unique challenges posed by Vision Transformers. Specifically, we aim to resolve the following questions: (1) How can memory consumption during the search process be reduced? (2) How can the unfair operation selection caused by softmax be mitigated? Furthermore, we adopt a progressive search strategy to minimise the significant disparity between the search and retraining phases.

In summary, our main contributions are threefold:
\begin{itemize}
    \item To the best of our knowledge, this is the \textbf{first attempt that applies a gradient-based DARTS approach} to Vision Transformer architecture search, extending the application scope of NAS into the Transformer domain. Achieving this required the novel definition of a suitable search space and encoding scheme for Transformers, ensuring compatibility with the DARTS logic. 

    \item We propose two innovative solutions—\textbf{Attention-based Partial Token Selection} and \textbf{Operation Fairness Regularization}—to address the critical challenges of memory consumption and softmax-induced unfairness in operation selection. These techniques make the search process more efficient and stable regarding memory usage and operation exploration.

    \item The architectures discovered by our method exhibit \textbf{conceptual differences} compared to existing ViT architectures and consistently outperform ViT-B/16 across three experimental datasets, achieving significant improvements in accuracy while also delivering reductions in parameter counts and FLOPs.
\end{itemize}

\section{Related Work}
\subsection{Vision Transformer}
The vanilla Vision Transformer (ViT) \cite{dosovitskiy2020image} effectively models global relationships within images by dividing the input image into patches, performing linear embedding, and incorporating positional encoding alongside multiple layers of multi-head self-attention. Specifically, the Vision Transformer first divides an input image $\mathbf{I} \in \mathbb{R}^{H \times W \times C}$ into $N$ non-overlapping patches, each of size $P \times P$. Here, $H$ and $W$ denote the height and width of the image, respectively, and $C$ represents the number of channels. Each patch $\mathbf{I}_i \in \mathbb{R}^{P \times P \times C}$ is then flattened into a vector (a.k.a. token) and mapped to a $D$-dimensional embedding vector $\mathbf{E}_i \in \mathbb{R}^D$ using a linear embedding layer.  

To encode spatial positional information, a positional encoding matrix $\mathbf{P} \in \mathbb{R}^{N \times D}$ is added to the patch embeddings $\mathbf{E} \in \mathbb{R}^{N \times D}$, where $N = \frac{H \times W}{P^2}$ is the total number of patches. This addition ensures that Transformer captures the order and spatial arrangement of the patches within the input image. 

The ViT architecture comprises multiple stacked encoder layers, including a Multi-head Self-Attention (MSA) mechanism and a Multi-Layer Perceptron (MLP). By design, MSA models long-range dependencies across patches, while MLP refines the representations within each encoder block. 

\textbf{Multi-head Self-Attention.} The MSA block captures features from different subspaces by computing multiple attention heads in parallel. Each attention head, denoted as \(\text{head}_h\), is computed using the attention mechanism, which takes as input the query (\(\mathbf{Q}\)), key (\(\mathbf{K}\)), and value (\(\mathbf{V}\)) matrices. Specifically, \(\mathbf{Q}\), \(\mathbf{K}\), and \(\mathbf{V}\) are obtained by projecting the input \(\mathbf{Z}_{l-1} \in \mathbb{R}^{N \times D}\) using learnable weight matrices \(\mathbf{W}_h^Q \in \mathbb{R}^{D \times d_k}\), \(\mathbf{W}_h^K \in \mathbb{R}^{D \times d_k}\), and \(\mathbf{W}_h^V \in \mathbb{R}^{D \times d_v}\), respectively:  
\begin{equation}
    \mathbf{Q} = \mathbf{Z}_{l-1} \mathbf{W}_h^Q, \quad \mathbf{K} = \mathbf{Z}_{l-1} \mathbf{W}_h^K, \quad \mathbf{V} = \mathbf{Z}_{l-1} \mathbf{W}_h^V
\end{equation}

The outputs of all heads are concatenated and projected using another learnable weight matrix \(\mathbf{W}^O \in \mathbb{R}^{(H \cdot d_v) \times D}\) to produce the final output of the MSA block. The attention operation for each head is computed as:  
\begin{equation}
    \text{Attention}(\mathbf{Q}, \mathbf{K}, \mathbf{V}) = \text{softmax}\left(\frac{\mathbf{Q} \mathbf{K}^\top}{\sqrt{d_k}}\right) \mathbf{V}
\end{equation}

where \(d_k\) is the dimensionality of the key vectors, and the softmax function ensures the attention scores are normalised.  

The overall process is expressed as:
\begin{equation}
    \text{MSA}(\mathbf{Z}_{l-1}) = \text{Concat}(\text{head}_1, \text{head}_2, \dots, \text{head}_H) \mathbf{W}^O
\end{equation}
where \(H\) is the number of attention heads.

\textbf{Multi-Layer Perceptron.} The MLP block applies two linear transformations with a non-linear activation function (typically GELU) in between, enabling position-wise feature transformations. The computation is given by:  
\begin{equation}
    \text{MLP}(\mathbf{Z}) = \text{GELU}(\mathbf{Z} \mathbf{W}_1 + \mathbf{b}_1) \mathbf{W}_2 + \mathbf{b}_2
\end{equation}
where \(\mathbf{W}_1 \in \mathbb{R}^{D \times D_h}\), \(\mathbf{W}_2 \in \mathbb{R}^{D_h \times D}\), and \(\mathbf{b}_1, \mathbf{b}_2\) are learnable parameters. A special classification token (\texttt{[CLS]}) is prepended to the input sequence, and its corresponding output vector is used for the final classification task.  

Unlike traditional Convolutional Neural Networks (CNNs), ViT does not rely on local convolutional operations. Instead, it leverages a global self-attention mechanism to model long-range dependencies across the entire image within the sequence space. With sufficient training data or pre-training on large-scale datasets (e.g., ImageNet-21k or JFT), ViT can match or surpass CNNs in performance. Moreover, its parallelizable and straightforward structure allows for efficient scaling to larger models, achieving strong performance across diverse vision tasks. Subsequent research has improved ViT in three key areas: enhancing data efficiency through distillation and more substantial data augmentation and regularization, reducing reliance on large datasets \cite{touvron2021training}; introducing architectural innovations such as multi-scale feature modelling and local-window attention mechanisms \cite{liu2021swin, han2021transformer, chu2021twins}, which enhance hierarchical and fine-grained feature extraction; and improving training stability and scalability \cite{touvron2021going, heo2021rethinking, wang2021pyramid}.

\begin{figure*}[!t]
\centering
\includegraphics[width=\linewidth]{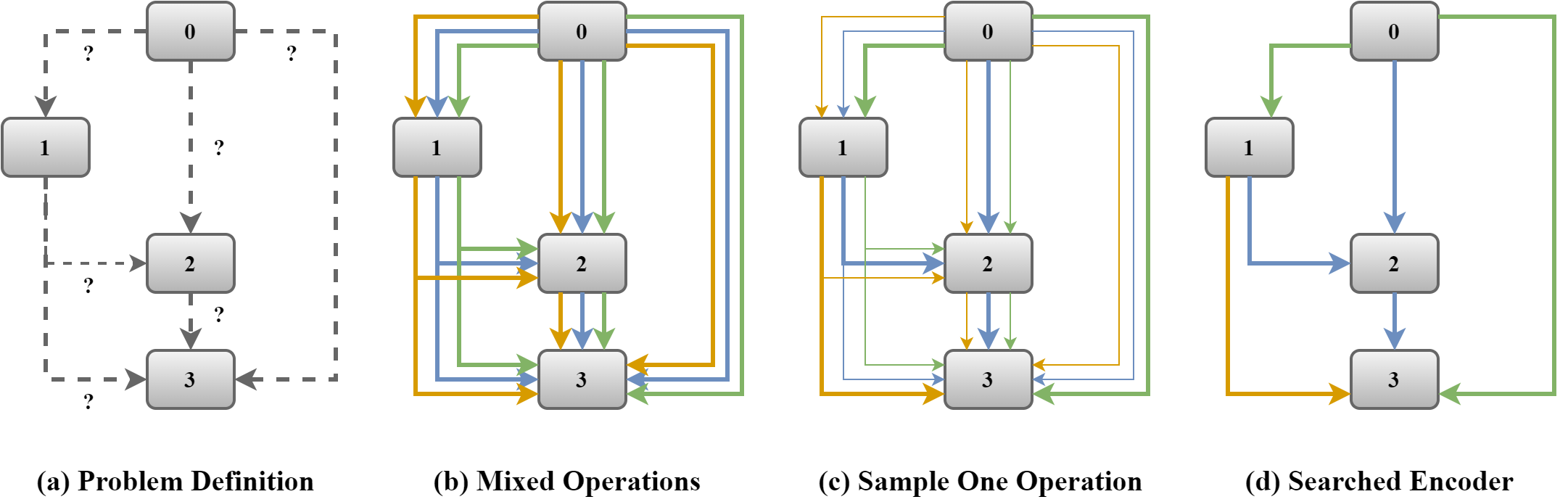}
\caption{An overview of DASViT. Nodes represent token embeddings (feature vectors) within each Transformer Encoder, and edges denote specific transformations selected from candidate operations.}
\label{fig:DASViT}
\end{figure*}

\subsection{Differentiable Architecture Search}
DARTS \cite{liu2018darts} reformulates the traditional discrete architecture search problem into a continuous optimisation problem, enabling efficient automated neural network design. Specifically, DARTS parameterises a set of candidate operations at the cell level. Given a set of candidate operations \(\mathcal{O}\) (e.g., convolution, pooling, or skip connections), each potential edge \(m\) in the cell is associated with a group of learnable parameters \(\boldsymbol{\alpha}^{(m)} = \{\alpha_o^{(m)} \mid o \in \mathcal{O}\}\).  

During the forward pass, the output of edge \(m\) is represented as a weighted sum of all candidate operations, where the weights are determined by a softmax function over \(\boldsymbol{\alpha}^{(m)}\):  
\begin{equation}
    o^{(m)}(\mathbf{x}) = \sum_{o \in \mathcal{O}} \frac{\exp(\alpha_o^{(m)})}{\sum_{o' \in \mathcal{O}} \exp(\alpha_{o'}^{(m)})} o(\mathbf{x})
\end{equation}
where \(o^{(m)}(\mathbf{x})\) is the output of edge \(m\) for input \(\mathbf{x}\), and \(\alpha_o^{(m)}\) represents the learnable weight for operation \(o\) on edge \(m\). The softmax operation ensures the weights are normalised, effectively acting as a probabilistic mixture of operations.

This formulation enables the architecture parameters \(\boldsymbol{\alpha}\) to be optimised jointly with the model weights \(\boldsymbol{\theta}\) using gradient-based methods. During the optimisation phase, DARTS alternates updates between the training and validation sets, guiding the architecture towards configurations that generalise better on the validation set.  

Once the search process is complete, the learned architecture parameters \(\boldsymbol{\alpha}\) are discretised by selecting the operation with the highest weight on each edge. This process yields a definitive network structure. By leveraging gradient information to update \(\boldsymbol{\alpha}\) and integrating the search process with model training, DARTS achieves significant speedup and reduced resource consumption compared to traditional methods based on reinforcement learning or evolutionary algorithms.  

Despite its efficiency, DARTS faces challenges such as overfitting architecture parameters and convergence to degenerate solutions (e.g., an overreliance on skip connections). To address these issues, subsequent research~\cite{xu2019pc, chen2019progressive, liang2019darts+, chu2020darts, chu2020fair, ye2022b} has proposed various improvements, including enhanced search stability, reduced computational costs, and improved generalisation of the final architectures.

\subsection{Architecture Search on Vision Transformer}
Since the introduction of ViT, numerous studies have sought to automate network architecture search for this model. In the domain of pure self-attention ViT architectures (as opposed to hybrid models combining convolutions), Chen et al.~\cite{chen2021autoformer} proposed a weight-entangled one-shot NAS combined with evolutionary search, incorporating key Transformer dimensions—such as attention heads, embedding size, Q/K/V parameters, MLP ratio, and depth—into the search space. To further refine the ViT search process, Chen et al.\cite{chen2021searching} introduced a two-step strategy: first optimising the search space, then training a supernet and conducting an evolutionary search under constraints.

Building on these ideas, Su et al.\cite{su2022vitas} highlighted the limitations of traditional sequential weight-sharing mechanisms. To address this, they introduced a cyclical weight-sharing strategy alongside Identity Shifting, which uses identity operations at deeper layers to reduce redundancy and improve search efficiency and accuracy. They also introduced separate class tokens for different patch sizes to mitigate path interference. Ni et al.\cite{ni2021nasformer} proposed an Expansion Window Self-Attention mechanism inspired by dilated convolutions to capture global dependencies better. They integrated this mechanism into their ViT architecture search process.

These studies focus on optimising key network dimensions and following a similar procedure: training a supernet and then using evolutionary algorithms to identify the best-performing subnet. In contrast, our research aims to go beyond fine-tuning existing dimensions by discovering entirely new architectures within the Transformer encoder layer.

\section{Methodology: The proposed Differentiable Architecture Search for Vision Transformer}
\subsection{Overview of the Method}
Our proposed \textbf{D}ifferentiable \textbf{A}rchitecture \textbf{S}earch for \textbf{Vi}sion \textbf{T}ransformer (DASViT) builds upon ViT-B/16 \cite{dosovitskiy2020image} as the baseline, aiming to search for the topology of the Transformer Encoder. This encoder consists of stacked MSA and MLP operations, forming a deep, modular network for image classification tasks.

The optimisation objective of DASViT is formulated as a bi-level problem:
\begin{equation}
    \min_{\boldsymbol{\alpha}} \left( \mathcal{L}_{val} + \mathcal{L}_{fair} \right)(w^*(\boldsymbol{\alpha}), \boldsymbol{\alpha})
\end{equation}
\begin{equation}
    \text{s.t.} \quad w^*(\boldsymbol{\alpha}) = \arg \min_{w} \mathcal{L}_{train}(w, \boldsymbol{\alpha})
\end{equation}
where $\mathcal{L}_{val}$ and $\mathcal{L}_{train}$ represent the losses on the validation and training sets, respectively. The upper-level optimisation aims to find the architecture parameters $\boldsymbol{\alpha}$ that minimise the sum of $\mathcal{L}_{val}$ and the operation fairness regularisation term $\mathcal{L}_{fair}$. At the same time, the lower-level problem optimises the model weights $w$ to minimise $\mathcal{L}_{train}$.

As illustrated in Figure~\ref{fig:DASViT} (a), the search problem is framed as the selection of appropriate edges in a Directed Acyclic Graph (DAG) with multiple nodes, where each edge represents an operation chosen from a set of candidate operations. Each unit consists of \( N \) nodes and for each pair of nodes \((i, j)\) (where \( i < j \)), an edge \((i, j)\) connects them, requiring an operation to be selected from the candidate set. The set of operations \( \mathcal{O} = \{ O_k \}_{k=1}^K \) includes fundamental functions such as MSA and MLP.

To make the search space continuous, as shown in Figure~\ref{fig:DASViT} (b), a mixed-edge representation is employed, where the outputs of all candidate operations are combined as a weighted sum. Specifically, for each edge \((i, j)\), the combined output of candidate operations is given by:
\begin{equation}
    \bar{o}_{i,j}(x_i) = \sum_{o_k \in \mathcal{O}} \frac{\exp(\alpha_{i,j}^k)}{\sum_{o_{k'} \in \mathcal{O}} \exp(\alpha_{i,j}^{k'})} \cdot o_k(x_i)
\end{equation}
where \( \alpha_{i,j}^k \) represents the weight for operation \( O_k \) on edge \((i, j)\). The softmax function transforms the discrete operation selection into a continuous weight distribution, optimising the architecture search via gradient descent. Data flows between nodes based on the weighted operations along the edges. Each node \( j \) receives inputs from its preceding nodes, represented as a weighted sum of the outputs from edges \((i, j)\) where \( i < j \):
\begin{equation}
    x_j = \sum_{i < j} \bar{o}_{i,j}(x_i)
\end{equation}

The specific data flow in the DAG structure is as follows:
\begin{equation}
    x_1 = \bar{o}_{0,1}(x_0)
\end{equation}
\begin{equation}
    x_2 = \bar{o}_{0,2}(x_0) + \bar{o}_{1,2}(x_1)
\end{equation}
\begin{equation}
    x_3 = \bar{o}_{0,3}(x_0) + \bar{o}_{1,3}(x_1) + \bar{o}_{2,3}(x_2)
\end{equation}

Finally, as shown in Figure~\ref{fig:DASViT} (c) and (d), the search process optimises the operation weights for each edge. The resulting encoder structure retains only the operations with the highest weights.

\begin{algorithm}[htbp]
    \caption{Framework of DASViT}
    \label{alg:DASViT}
    \begin{algorithmic}[1]
        \Require Number of stages \( N \); number of epochs per stage \( E \); number of network layers \( L_n \); set of candidate operations \( \mathcal{O}_n \); number of operations to prune per stage \( P_n \)
        \Ensure The optimal candidate architecture \( \boldsymbol{\alpha}^* \)
        \State Initialize architecture parameters \( \boldsymbol{\alpha} \) and network weights \( \boldsymbol{w} \)
        \State \( i \gets 1 \)
        \For{$i \leq N$}
            \State Set the number of layers \( L_n = L_1 + (n-1) \times \Delta L \)
            \State Build the network with \( L_n \) layers using candidate operations \( \mathcal{O}_n \)
            \If{$i > 1$}
                \State Inherit network weights from the previous stage \( i - 1 \)
            \EndIf
            \State \( j \gets 1 \)
            \For{$j \leq E$}
                \State Create the mixed operation parametrized by \( \boldsymbol{\alpha} \) with top-\(k\) tokens selected
                \State Compute regularization loss \( \mathcal{L}_{fair} \) based on \( \boldsymbol{\alpha} \)
                \State optimise architecture \( \boldsymbol{\alpha} \) via
                \[
                \nabla_{\boldsymbol{\alpha}} \left( \mathcal{L}_{val} + \mathcal{L}_{fair} \right)\left(\boldsymbol{w}^*(\boldsymbol{\alpha}) - \xi \nabla_{\boldsymbol{w}} \mathcal{L}_{train}(\boldsymbol{w}, \boldsymbol{\alpha}), \boldsymbol{\alpha} \right)
                \]
                \State optimise weights \( \boldsymbol{w} \) via \( \nabla_{\boldsymbol{w}} \mathcal{L}_{train}(\boldsymbol{w}, \boldsymbol{\alpha}) \)
                \State \( j \gets j + 1 \)
            \EndFor
            \State Prune the \( P_n \) least promising operations from \( \mathcal{O}_n \) to obtain \( \mathcal{O}_{n+1} \)
            \State Retain the top \( |\mathcal{O}_{n+1}| \) candidate operations for the next stage
            \State \( i \gets i + 1 \)
        \EndFor
        \State Derive the optimal architecture \( \boldsymbol{\alpha}^* \) based on the learned \( \boldsymbol{\alpha} \)
        \State \Return The optimal candidate architecture \( \boldsymbol{\alpha}^* \)
    \end{algorithmic}
\end{algorithm}

To reduce memory consumption, we propose attention-based partial token selection, which selects the top-$k$ tokens for training. To address the network depth gap between search and retraining phases \cite{chen2019progressive}, the algorithm is divided into $N$ stages. Initially, all candidate operations are retained, and the network depth is set to 2 layers. Underperforming operations are pruned at the end of each stage based on their weights. The top five operations are retained in intermediate stages, and the network depth increases to 4 layers. The top three operations are retained in the final stage, and the depth expands to 6 layers. The overall algorithm is summarized in Algorithm~\ref{alg:DASViT}.

\subsection{Search Space}\label{subsec:search_space}
The search space design is based on the vanilla ViT encoder, which consists of MSA and MLP operations. To ensure fairness in selecting different types of operations, the search space maintains a balanced ratio between the number of MSA and MLP operations. Moreover, we introduce various configurations for MSA and MLP, including variations in the number of attention heads and the hidden layer dimensions. To enhance the flexibility of the search space and improve the model's expressive capacity, we also incorporate two additional operations: the \texttt{none} (Zero) operation and the \texttt{skip-connect} (Identity) operation. The \texttt{none} operation eliminates unnecessary connections, simplifying the network structure and reducing computational complexity. In contrast, the \texttt{skip-connect} operation improves gradient flow and facilitates feature retention by directly passing the input.

Specifically, the search space includes eight operations:

\begin{itemize}
    \item \textbf{Zero}: Outputs a zero tensor, effectively removing unnecessary connections.
    \item \textbf{Identity}: Passes the input directly, implementing skip-connect in the graph.
    \item \textbf{MSA}: Multi-head Self-Attention, with the number of attention heads \( h \in \{8, 12, 16\} \).
    \item \textbf{MLP}: Multi-Layer Perceptron, with MLP ratio \( r \in \{0.5, 3, 4\} \).
\end{itemize}

Here, the MLP ratio refers to the ratio of the hidden layer dimension to the input feature dimension in the MLP.

\subsection{Attention-based Partial Token Selection}
To reduce memory consumption during the search stage, we propose an attention-based partial token selection mechanism to focus on the most informative parts of the input data. Let the input token sequence be represented as a tensor \( \mathbf{X} \in \mathbb{R}^{B \times N \times C} \), where \( B \) is the batch size, \( N \) is the number of tokens, and \( C \) is the embedding dimension. First, the queries (\(\mathbf{Q}\)) and keys (\(\mathbf{K}\)) are obtained through linear transformations:
\begin{equation}
    \mathbf{Q} = \mathbf{X} \mathbf{W}_Q, \quad \mathbf{K} = \mathbf{X} \mathbf{W}_K
\end{equation}
where \( \mathbf{W}_Q, \mathbf{W}_K \in \mathbb{R}^{C \times C} \) are trainable weight matrices. Subsequently, the attention scores are computed using the scaled dot-product attention mechanism:
\begin{equation}
    \text{score}_{i,j} = \frac{\mathbf{Q}_i \mathbf{K}_j^\top}{\sqrt{C}}, \quad \forall i, j \in \{1, \dots, N\}
\end{equation}

For each token \(i\), the comprehensive attention score is obtained by averaging all scores:
\begin{equation}
    \mathbf{s}_i = \frac{1}{N} \sum_{j=1}^{N} \text{score}_{i,j}
\end{equation}
Based on the scores, the top \( k = \lfloor \lambda N \rfloor \) tokens with the highest scores are selected, where \( \lambda \in (0,1] \) is the selection proportion parameter. This selection process can be formulated as \( \mathcal{I} = \text{TopK}(\mathbf{s}, k) \), where \( \mathcal{I} \) contains the indices of the selected tokens. The resulting subset of tokens is then given by \( \mathbf{X}_{\mathbf{s}} = \mathbf{X}[\mathcal{I}] \).

At the initial training phase, the weight matrices \( \mathbf{W}_Q \) and \( \mathbf{W}_K \) are randomly initialised, causing the initial attention scores \( \mathbf{s}_i \) to reflect the importance of tokens potentially inaccurately. However, as training progresses, \( \mathbf{W}_Q \) and \( \mathbf{W}_K \) are optimised through backpropagation and optimiser updates, resulting in more discriminative attention scores. Consequently, the selected token subset \( \mathbf{X}_{\mathbf{s}} \) increasingly tends to include the tokens that contribute more significantly to the task (classification), effectively improving the selection accuracy and overall performance of the model.

\subsection{Operation Fairness Regularization}
Operation fairness regularization consists of two components, each addressing the concerns about fairness between skip-connect operations and other operation types, respectively. By introducing fairness constraints, it aims to balance the selection probabilities of different operations, ensuring that no single operation type dominates the architecture search process. This approach guarantees that each operation has a fair opportunity to participate in the competition.

\textbf{Fairness Regularization for Skip Connections.} To prevent the skip-connect operation from being disproportionately favoured during the optimisation process, we define the set of skip-connect operations as \( \mathcal{O}_{\text{skip}} \subseteq \mathcal{O} \). The fairness regularisation term \( \mathcal{L}_1 \) is computed as the average of the skip-connect operation weights:
\begin{equation}
    \mathcal{L}_1 = \frac{1}{|O_{\text{skip}}|} \sum_{o \in O_{\text{skip}}} \alpha_o
\end{equation}

Using the average rather than the sum operation, we ensure that the regularization term \( \mathcal{L}_1 \) remains independent of the number of skip-connect operations. This approach effectively controls the average selection probability of skip-connect relative to all operations, preventing it from dominating the architecture search process.

\textbf{Fairness Regularization for Operation Types.} To balance the weight distribution among different operation types (e.g., MLP and MSA), the operation set \( \mathcal{O} \) is partitioned into distinct types \( T = \{T_1, T_2, \dots, T_K\} \). We impose constraints on the total weights of each type separately. When the total weight of a particular type exceeds the predefined upper limit \( \gamma_{\text{max}} \) or falls below the lower limit \( \gamma_{\text{min}} \), corresponding penalties or rewards are applied to maintain a balanced distribution of operation types. For each operation type \( T_k \), the total weight is defined as:
\begin{equation}
    \alpha_{T_k} = \sum_{o \in \mathcal{O}_{T_k}} \alpha_o
\end{equation}
We set lower and upper threshold values \( \gamma_{\text{min}} \) and \( \gamma_{\text{max}} \). When the total weight \( \alpha_{T_k} \) of a type \( T_k \) exceeds this range, we apply a penalty or reward accordingly. The fairness regularisation term for different operation types \( \mathcal{L}_2 \) is defined as:
\begin{equation}
    \mathcal{L}_2 = \sum_{k=1}^{K} \left[ \zeta_1 \cdot \max \left( 0, \alpha_{T_k} - \gamma_{\max} \right) + \zeta_2 \cdot \max \left( 0, \gamma_{\min} - \alpha_{T_k} \right) \right]
\end{equation}
where \( \zeta_1 \) is the penalty coefficient applied when exceeding the upper limit, and \( \zeta_2 \) is the reward coefficient applied when falling below the lower limit.

\textbf{Overall Fairness Regularization.} Combining the fairness regularization for skip-connect and different operation types, the total fairness regularization term \( \mathcal{L}_{fair} \) is defined as:
\begin{equation}
    \mathcal{L}_{fair} = a \cdot \mathcal{L}_1 + b \cdot \mathcal{L}_2
\end{equation}
where \( a \)  and \( b \) are balancing parameters that control the contribution of each fairness item to the total loss.

\section{Experiments}
\begin{table*}[htbp]
\caption{Comparison of Top-1 and Top-5 Accuracy across different datasets and models without pre-training.}
\centering
\begin{tabular}{c|cc|cc|cc|c|c|c}
\hline
\multirow{2}{*}{Model} & \multicolumn{2}{c|}{CIFAR-10} & \multicolumn{2}{c|}{CIFAR-100} & \multicolumn{2}{c|}{ImageNet-100} & \multirow{2}{*}{\#Params} & \multirow{2}{*}{FLOPs} & \multirow{2}{*}{Design Type} \\ \cline{2-7}
 & Top-1 Acc & Top-5 Acc & Top-1 Acc & Top-5 Acc & Top-1 Acc & Top-5 Acc & \multicolumn{1}{c|}{}&\multicolumn{1}{c|}{}\\ \hline
\multicolumn{10}{c}{Convolutional Neural Networks} \\ \hline
VGG-16 \cite{simonyan2014very} & 72.5\% & 97.1\% & 37.0\% & 61.5\% & 32.0\% & 55.7\% & 134.7M & 15.5G & Manual \\
VGG-19 \cite{simonyan2014very} & 73.1\% & 97.3\% & 37.2\% & 62.0\% & 30.3\% & 54.3\% & 140.0M & 19.7G & Manual \\ 
ResNet-50 \cite{he2016deep} & 81.5\% & 98.7\% & 52.4\% & 80.7\% & 59.8\% & 83.2\% & 23.7M & 4.1G & Manual \\ 
ResNet-152 \cite{he2016deep} & 80.2\% & 98.6\% & 53.3\% & 80.7\% & 57.8\% & 82.9\% & 58.4M & 11.6G & Manual \\ 
EfficietNet-B7 \cite{tan2019efficientnet} & 84.4\% & 98.8\% & 56.7\% & 81.9\% & 53.4\% & 77.6\% & 64.0M & 5.2G & Auto \\
RegNetY-12GF \cite{radosavovic2020designing} & 82.3\% & 99.0\% & 54.4\% & 81.8\% & 57.9\% & 82.7\% & 49.8M & 12.1G & Auto \\ 
RegNetY-16GF \cite{radosavovic2020designing} & 82.6\% & 98.8\% & 54.4\% & 81.6\% & 58.7\% & 83.2\% & 80.7M & 15.9G & Auto \\ \hline
\multicolumn{10}{c}{Transformers} \\ \hline
ViT-B/16 \cite{dosovitskiy2020image} & 78.8\% & 97.9\% & 45.7\% & 72.0\% & 39.7\% & 66.1\% & 85.8M & 12.0G & Manual \\
TNT-B \cite{han2021transformer} & 76.1\% & 94.2\% & 48.0\% & 75.1\% & 45.6\% & 68.6\% & 64.8M & 13.5G & Manual \\
CaiT-S36 \cite{touvron2021going} & 78.6\% & 97.5\% & 46.6\% & 72.7\% & 39.3\% & 64.6\% & 67.9M & 12.9G & Manual \\
AutoFormer-B \cite{chen2021autoformer} & 79.2\% & 97.8\% & 47.1\% & 73.4\% & 41.2\% & 66.9\% & 52.5M & 11.6G & Auto \\
\textbf{DASViT (ours)} & \textbf{80.1\%} & \textbf{98.4\%} & \textbf{54.4\%} & \textbf{78.5\%} & \textbf{46.8\%} & \textbf{72.3\%} & \textbf{50.4M} & \textbf{9.9G} & Auto \\ \hline
\end{tabular}
\label{tab:model_comparison}
\end{table*}

\subsection{Experimental Setup}
In this study, we first conduct an architecture search on CIFAR-10 and then apply the best-found architecture to both CIFAR-10 and CIFAR-100 for retraining. This approach leverages the similarity between the datasets regarding class count and image characteristics to assess the architecture’s adaptability and performance across related datasets. Additionally, we perform an independent architecture search on ImageNet-100 and retrain it to explore performance on a larger, more diverse dataset. All images are resized to 224×224 pixels.

The CIFAR-10 dataset \cite{krizhevsky2009learning} contains 60,000 images across 10 categories, with 50,000 used for training and 10,000 for testing. CIFAR-100 \cite{krizhevsky2009learning} extends CIFAR-10 with 100 fine-grained categories, each containing 600 images. It consists of 60,000 images, with 50,000 for training and 10,000 for testing. ImageNet-100 is a subset of the ImageNet dataset \cite{deng2009imagenet}, containing 100 categories with approximately 1,000 images per category. It includes 105,771 training and 5,000 validation images, sampled according to \cite{tian2020contrastive}.

The algorithm is implemented in PyTorch, and all experiments are conducted on a single NVIDIA GeForce RTX 3090 GPU with 24GB of memory. The baseline model is ViT-B/16 \cite{dosovitskiy2020image}, with an embedding dimension of 768, an input image size of $224 \times 224$, and a patch size of 16. The search and retraining phases are as follows:

\textbf{Search Phase.} The search process uses a progressive strategy, divided into three stages of 30 epochs each, for 90 epochs. The batch size is 64, with an initial learning rate of $1 \times 10^{-3}$ and a weight decay of $5 \times 10^{-2}$. AdamW is the optimiser, and the learning rate is adjusted using a Cosine Annealing scheduler.

\textbf{Retraining Phase.} The model is retrained for 500 epochs, with a warmup for the first 20, starting at a learning rate of $1 \times 10^{-6}$. The batch size is increased to 128 while the learning rate and weight decay remain unchanged. The optimiser and scheduler are identical to those used in the search phase.

For fairness, all the models are trained from scratch, without pre-training, and the experimental settings strictly follow those reported in the original papers.

\subsection{Search Results}
\begin{figure}
    \centering
    \includegraphics[width=0.9\linewidth]{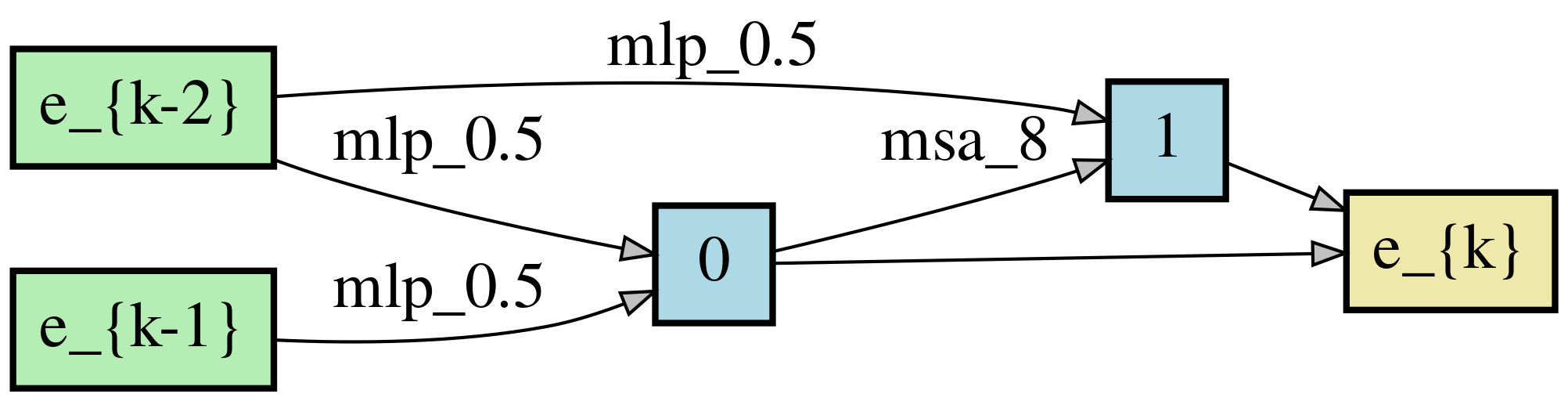}
    \caption{Transformer Encoder searched on CIFAR-10.}
    \label{fig:CIFAR}
\end{figure}

\begin{figure}
    \centering
    \includegraphics[width=0.9\linewidth]{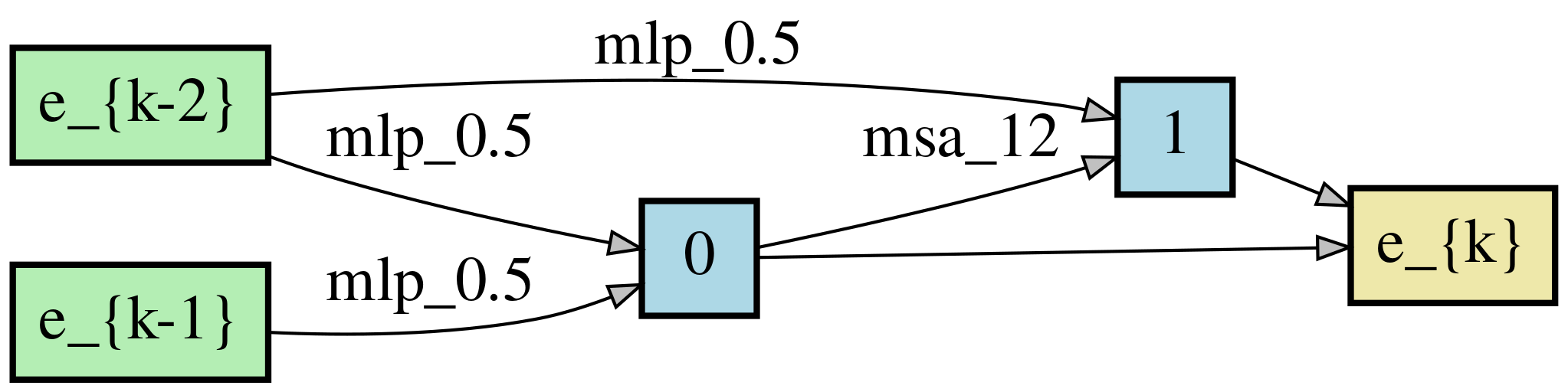}
    \caption{Transformer Encoder searched on ImageNet-100.}
    \label{fig:IMNT}
\end{figure}

Figure~\ref{fig:CIFAR} and Figure~\ref{fig:IMNT} illustrate the Transformer encoder architectures searched by our proposed DASViT on the CIFAR-10 and ImageNet-100 datasets, respectively.

\textbf{Encoder Data Flow.} Within the encoder, the data flow can be summarized as follows:

\begin{enumerate}
    \item The encoder takes two input embeddings, $e_{k-2}$ and $e_{k-1}$, from the previous layers. Each input is processed through an MLP module with an MLP ratio of 0.5, producing intermediate outputs.
    \item The intermediate results are summed to compute $o_0$:
    \begin{equation}
        o_0 = \text{MLP}(e_{k-2}) + \text{MLP}(e_{k-1})
    \end{equation}
    \item Next, $o_0$ is passed through an MSA block, which has either 8 or 12 attention heads depending on the dataset, to produce $o_1$. Additionally, $e_{k-2}$ is processed again by an MLP module and added to $o_1$:
    \begin{equation}
        o_1 = \text{MSA}(o_0) + \text{MLP}(e_{k-2})
    \end{equation}
    \item Finally, the encoder output $e_k$ is computed as the sum of $o_0$ and $o_1$:
    \begin{equation}
        e_k = o_0 + o_1
    \end{equation}
\end{enumerate}

\textbf{Overview Workflow.} The complete workflow of the architecture can be summarized by the following equations:
\begin{align}
    \mathbf{E}_i &= \text{Linear}(\text{Flatten}(\mathbf{I}_i)) \quad \forall i \in \{1, 2, \ldots, N\} \\
    \mathbf{Z}_0 &= \mathbf{E}_i + \mathbf{P}_i\\
    \mathbf{Z}'_{\ell} &= \text{MLP}(\mathbf{Z}_{\ell-2}) + \text{MLP}(\mathbf{Z}_{\ell-1}), \quad \ell = 1, 2, \ldots, L \\
    \mathbf{Z}''_{\ell} &= \text{MSA}(\mathbf{Z}'_{\ell}) + \text{MLP}(\mathbf{Z}_{\ell-2}), \quad \ell = 1, 2, \ldots, L \\
    \mathbf{Z}_{\ell} &= \mathbf{Z}'_{\ell} + \mathbf{Z}''_{\ell}, \quad \ell = 1, 2, \ldots, L \\
    \mathbf{y} &= \text{MLP}(\mathbf{Z}_L[\texttt{CLS}])
\end{align}

For \( l = 1 \), both \( \mathbf{Z}_{l-2} \) and \( \mathbf{Z}_{l-1} \) default to the initial embedding \( \mathbf{Z}_0 \), which is derived from the input tokens and position embeddings.

\subsection{Retraining Results}
As shown in Table~\ref{tab:model_comparison}, DASViT outperforms ViT-B/16 on CIFAR-10 and CIFAR-100 in Top-1 and Top-5 accuracies. It reduces parameters by 41\% (50.4M vs 85.8M) and FLOPs from 12.0 to 9.9G. While pure self-attention models typically require large-scale pre-training and lag behind CNNs when trained from scratch, DASViT achieves comparable performance to efficient CNNs (e.g., ResNet-152, RegNetY-12GF) without pre-training while maintaining a significant FLOP advantage.

On ImageNet-100, DASViT surpasses ViT-B/16 but is about 15\% behind CNNs. This gap reflects CNNs' advantages in leveraging built-in priors like local receptive fields and translational invariance, enabling them to learn fine-grained features from increased data. In contrast, Transformers depend more on large-scale data to capture global relationships, making them more dataset-size and structure-sensitive.

\section{Conclusion and Future Work}
This study introduces a novel gradient-based NAS algorithm, named DASVIT, that modifies and adapts the logic of DARTS to design new ViT architectures.

The proposed DASViT encodes the search problem into a continuous space, enabling gradient-based optimisation techniques. By leveraging the efficiency of DARTS, traditionally used for CNNs, we discover novel ViT architectures. We introduce Attention-based Partial Token Selection and Operation Fairness Regularization to address challenges during the DARTS search process. Our progressive search strategy also bridges the gap between the search and retraining phases. The experimental results obtained on the CIFAR-10, CIFAR-100, and ImageNet-100 datasets demonstrate that the ViT architectures discovered with DASViT outperform ViT-B/16 in accuracy, parameter count, and FLOPs while narrowing the gap with CNNs even without pre-training. 

Future work should focus on developing a novel data augmentation technique and specialised large-scale pre-training strategy to unlock DASViT’s potential further and enhance its performance, as DASViT fundamentally diverges from the vanilla ViT architecture.

\section*{Acknowledgment}
This research was supported by the China Scholarship Council (No. 202308060192) and the Jiangsu Distinguished Professor Programme. 

\bibliographystyle{IEEEtran}
\bibliography{references}

\end{document}